# Computer Vision for Particle Size Analysis of Coarse-Grained Soils

*Sompote Youwai[1]  Parchya Makam[2]*

[1]*Associate Professor*
*Department of Civil Engineering, King Mongkut's University of Technology Thonburi*
[2]*Chief Executive Officer*
*Khonkaen Soil Engineering Company Limited*

**ABSTRACT:** Particle size analysis (PSA) is a fundamental technique for evaluating the physical characteristics of soils. However, traditional methods like sieving can be time-consuming and labor-intensive. In this study, we present a novel approach that utilizes computer vision (CV) and the Python programming language for PSA of coarse-grained soils, employing a standard mobile phone camera. By eliminating the need for a high-performance camera, our method offers convenience and cost savings. Our methodology involves using the OPENCV library to detect and measure soil particles in digital photographs taken under ordinary lighting conditions. For accurate particle size determination, a calibration target with known dimensions is placed on a plain paper alongside 20 different sand samples. The proposed method is compared with traditional sieve analysis and exhibits satisfactory performance for soil particles larger than 2 mm, with a mean absolute percent error (MAPE) of approximately 6%. However, particles smaller than 2 mm result in higher MAPE, reaching up to 60%. To address this limitation, we recommend using a higher-resolution camera to capture images of the smaller soil particles. Furthermore, we discuss the advantages, limitations, and potential future improvements of our method. Remarkably, the program can be executed on a mobile phone, providing immediate results without the need to send soil samples to a laboratory. This field-friendly feature makes our approach highly convenient for on-site usage, outside of a traditional laboratory setting. Ultimately, this novel method represents an initial disruption to the industry, enabling efficient particle size analysis of soil without the reliance on laboratory-based sieve analysis.

## 1. INTRODUCTION

Particle size distribution or gradation is a key parameter for characterizing soil behavior and classification(ASTM, 2009). Soil strength and stiffness depend largely on the particle size analysis of soil. The conventional method to determine the gradation of soil is to use sieve analysis in a laboratory. Sieve analysis involves separating soil particles by passing them through different sizes of mesh. Soil particles that are larger than the mesh size are retained on the sieve. The weight of soil retained on each sieve is used to calculate the particle distribution of soil. However, this method is time-consuming and requires sophisticated and expensive equipment. Therefore, it would be beneficial to develop a method that can obtain the gradation of soil in real time in the field and enhance the quality control of construction.

Computer vision (CV) or image analysis is a valuable technique used to estimate soil particle size accurately from digital images. The basis of this method lies in establishing a correlation between the number of pixels and the true length of the soil particles. Previous investigations have successfully employed CV in controlled laboratory conditions ((Andersson, 2010; Dipova, 2017; Igathinathane et al., 2009; Zheng & Hryciw, 2017). These studies involved conveying soil particles to a fixed camera using a belt conveyor and illuminating them with a constant light source. Notably, the results obtained through CV were comparable to those obtained from traditional sieve analysis. In an attempt to determine the gradation of coarse-grained soil in laboratory settings, a novel image analysis method was proposed (Azarafza et al., 2021). Nevertheless, these techniques suffer from various limitations, including inconvenience, time consumption, and high costs. Therefore, there is a pressing need for the development of a more practical approach capable of determining soil particle size distribution in the field, utilizing a common mobile phone and instantly generating the soil gradation curve. Such an innovation could significantly disrupt the geotechnical engineering field, enhancing supervision and design practices.

This paper proposes a computer vision workflow to obtain the particle size analysis of soil. The soil can be placed on a simple A4 paper with a reference target. The target is used as a reference to determine the pixel-to-length ratio. The picture is taken by an ordinary digital camera in ambient light. The computer code of this study was develop by using Python (Van Rossum & Drake Jr, 1995)with open source liberally- OpenCV (Bradski, 2000; *OpenCV: OpenCV Modules*, n.d.). The development code was run on Google Colab. The python code developed in this study can be easily adapted to run as an application on mobile phones or web application. This paper will present the comparison of the results obtained from the code with those from sieve analysis. The limitations of the current approach and the future directions for improvement will also be discussed.

## 2. TEST MATERIAL

The 52 sand samples were obtained from a soil testing company affiliated with the second author and analyzed by dry sieving. The selection criterion was that the fraction of material finer than 0.075 mm (sieve No. 200) should be lower than 5%. The proposed COV program could not detect particles smaller than this limit due to the limited resolution of the camera. A higher resolution camera would enable the detection of finer soil particles. The details and discussion of the grain size distribution characteristics with different shapes and fine content are presented in the following section.

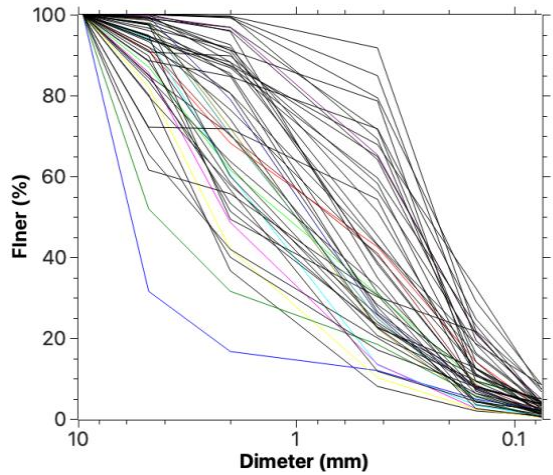

Fig. 1 The grain size distribution of sand

## 3. PROGRAM DEVELOPMENT

The way to estimate the size of an object using computer vision is to use a reference object with a known length. For example, a ruler can be placed near the object whose dimension is desired. The pixel-to-length ratio can be calculated by the following equation:

$$P = \frac{LP}{actual\ length} \quad (1)$$

Where $P$ is the pixel-to-length ratio, $LP$ is the length measured in pixels in the picture, and actual length is the actual length of the object. In this study, the unit of mm is used for the length of the object. Then, the actual length of the object can be estimated by the following equation:

$$length = P.LPP \quad (2)$$

where LPP is the length of the object in the unit of pixel. However, the manual calculates the pixel-to length manfully is time consuming.

This paper use the application of ArUco target (Garrido-Jurado et al., 2014; *OpenCV: Detection of ArUco Markers*, n.d.). An ArUco marker is a square synthetic marker that consists of a wide black border and an inner binary matrix that determines its identifier. The black border enables its fast detection in the image, and the binary code allows its identification and the application of error detection and correction techniques. ArUco markers are used for camera pose estimation in applications such as augmented reality and robotics. The ArUco marker can also detect the pixels at the corners of the marker. The coordinates of the corners of the marker can be used to calculate the perimeter of the marker. The first step to use this concept is to print an A4 paper with an ArUco marker on it (Fig. 2). The size of the maker could be use in different size depending on the size of the paper and the resolution of the camera. In this study, we use a marker with a width of 20 mm. Thus, the perimeter is 80 mm. The pixel-to-length ratio of the system can be calculated by Equation 1. It will be used to calculate the size of the size of the particle by using Equation 2.

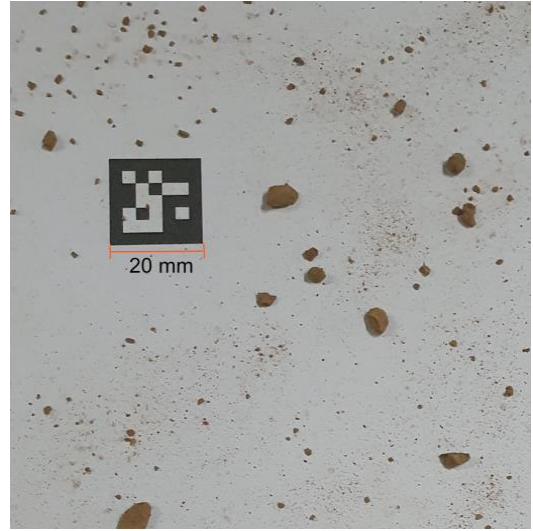

Fig.2 ArUco target and soil particle on A4 paper

The work flow of the particle size analysis program is shown in Fig. 3. The size of the particle was detected by using the available library in OpenCV. Firstly, the picture was detected by OpenCV and converted into a NumPy array (Harris et al., 2020) with RGB channel. Then, the picture was changed into grayscale to reduce the dimension of the matrix by 3 times compared to the array of the color image. The color in grayscale can be represented by a number from 0 to 255. The number 0 is black and 255 is white. Then, we tried to reduce noise in the image by using the Gaussian blur method. It is a technique that applies a linear filter to an image that smooths out noise and reduces detail (Szeliski, 2022). The filter uses a Gaussian kernel, which is a matrix of weights that depends on the distance from the center pixel. The filter convolves each pixel in the image with the kernel and then sums them up to produce the output pixel value. The values of the parameters are shown in Table 1. The parameters actually depend on the camera and lighting condition. The best parameters from this research were obtained by trial and error to achieve the best results.

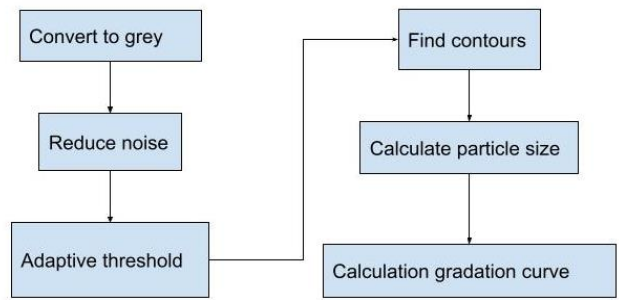

Figure 3 Workflow for gradation analysis of soil

Adaptive threshold technique (Roy et al., 2014; Ting, 2017) was applied to image to obtain the edge of the soil particle. It also use to handle images with non-uniform lighting conditions or low contrast. Adaptive thresholding is a technique that segments an image into foreground and background pixels by applying a local threshold value to each pixel. Unlike global thresholding methods that use a single threshold value for the entire image, adaptive thresholding computes the optimal threshold value for each region of the image based on its local characteristics, such as mean or Gaussian-weighted sum of pixel intensities. Threshold value of each pixel can be calculated from the local pixel from the following equation.

$$T(x,y) = \frac{1}{N}\sum_{i,j \in N(x,y)} I(i,j) - c \quad (3)$$

where $T(x,y)$ is the threshold value for pixel $(x,y)$, $N$ is the neighborhood $(x,y)$, $I(I,j)$ is the intensity value of pixel $(i,j)$ and $C$ is a constant. The example of the picture when apply Adaptive threshold technique is shown in Fig. 3 The size of the region to use for calculating the threshold value. In this study, we used a region size of 451 and a constant C of 72, as shown in Table 1. The region size determines how many pixels are used to calculate the threshold value. A larger region size makes the threshold value more robust to illumination changes, but also slows down the thresholding operation. The constant C adjusts the threshold value. A larger C makes the threshold value more conservative, while a smaller C makes the threshold value more aggressive.

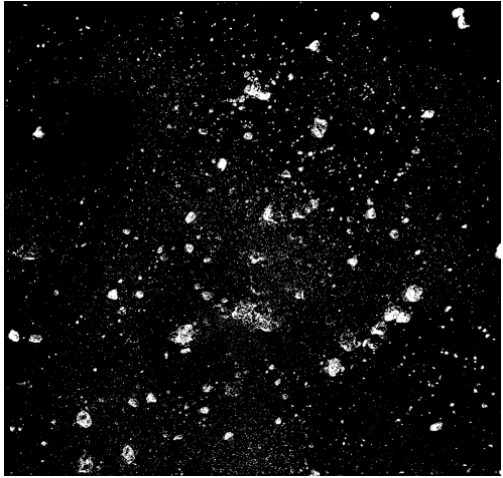

Fig. 3 The picture after process with adaptive threshold method

Table 1 Parameter for obtain the particle size of soil

| Function | parameters |
|---|---|
| Gaussian blur | ksize=(5,5) sigmax=10 |
| Adaptive thresholding | Block size =301 C= 72 |
| Find Contour | RETR_EXTERNAL CHAIN_APPROX_NONE |

To measure the soil particle size from an image, we use the find contour function in openCV(Bradski, 2000). This function detects and draws the boundaries of objects in a binary image. It returns a list of contours, which are curves that connect all the continuous points along an object's boundary with the same color or intensity. It also returns a hierarchy that describes the relationship between the contours. We set the find contour option to RETR_EXTERNAL, which means that only the outermost contours of objects are found, ignoring any holes inside them. This way, we can find only the external contour of the soil in the image.

We used the minAreaRect function in OpenCV (*OpenCV: Contour Features*, n.d.)to obtain the particle size from the previous contour. This function finds the minimum area rotated rectangle that encloses a set of 2D points and returns a Box2D structure with the center, width, height and angle of rotation of the rectangle. Figure 4 shows an example of the rectangle over a soil particle. We calculated the particle size as the average of the width and height of the rectangle. Then, we converted the length from pixel unit to mm unit using the pixel-to-length ratio from equation 1. Finally, we created the gradation soil curve based on the summation of the area of each particle.

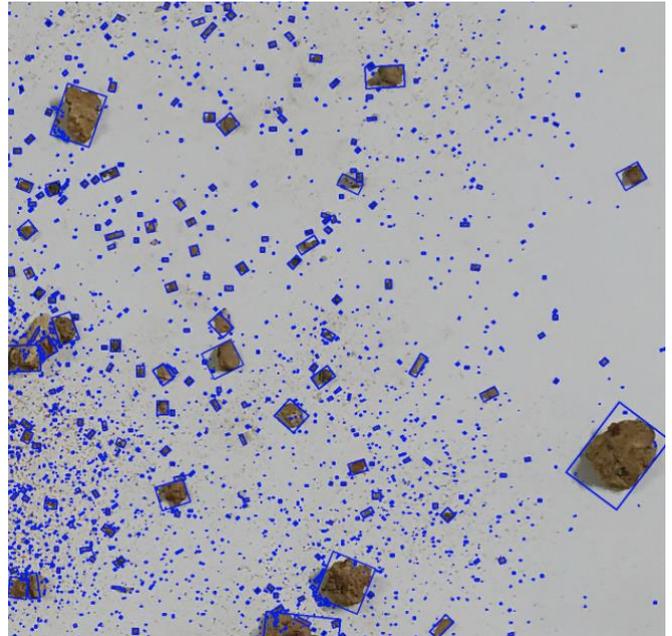

Fig. 4 The application of using liberally min area rectangular to obtain size of particle.

## 4. RESULT AND DISCUSSION

We placed the soil sample on A4 paper with an ARUCO mark on it (Fig. 5). This approach is very convenience in field. Gradation of soil can be get immediately without any special equipment. It is require only high resolution camera or mobile phone Each picture had a resolution of 3024x4032 pixels, equivalent to 12 megapixels. The pixel-to-length ratio was 14.2 pixels per millimeter. This implied that the system's high resolution was the inverse of the pixel-to-length coefficient, or 0.07 mm. Theoretically, this method could not detect soil particles smaller than 0.07 mm.

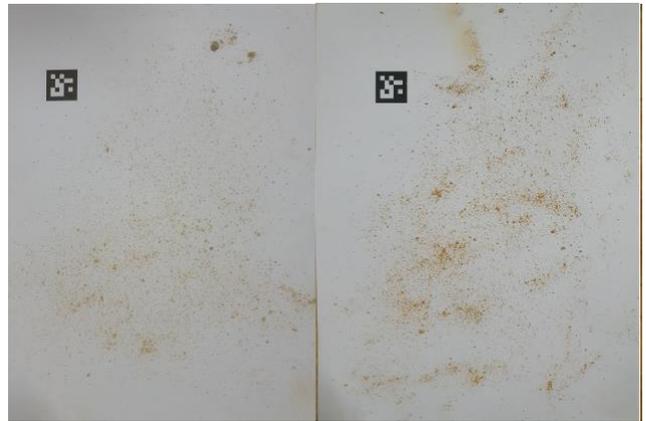

Fig. 5 example of soil particle on paper with ArUco targe

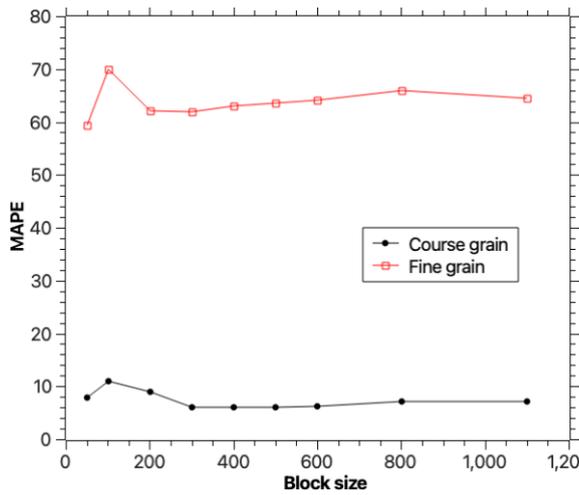

Fig. 6 The value of MAPE according to changing of block size

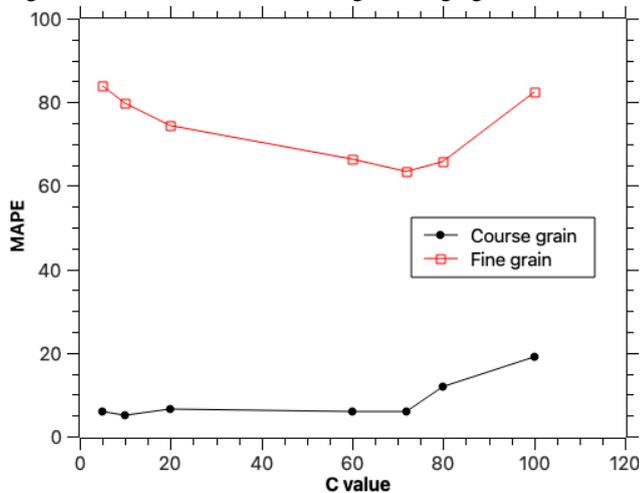

Fig. 7 The value of MAPE according to changing of C value

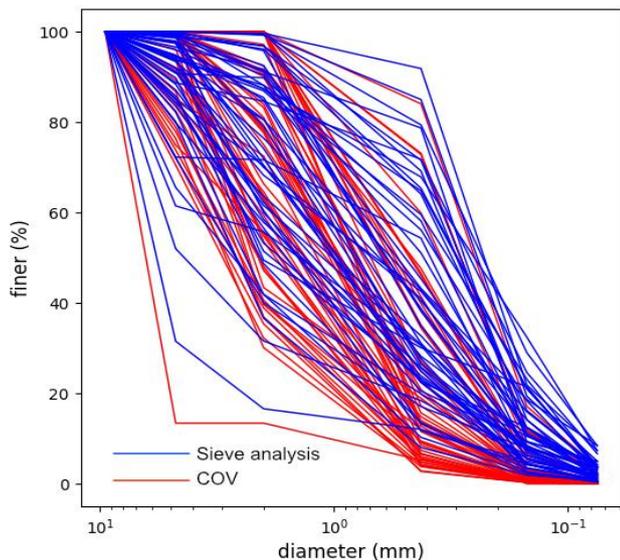

Fig. 8 The comparison between soil gradation from COV and sieve analysis

Table 2 mean square absolute error between COV and sieve analysis

| Particle size | Mean square absolute error (MAPE) |
|---|---|
| Coarse grain 2 -9.5 mm | 6.4% |
| Fine grain 0.075-0.425 | 62 % |

To achieve the lowest MAPE, we varied the parameter from the adaptive threshold library. We used c=71 and block size =301 as the initial values for the parametric study. Figs. 6 and 7 show the results of the parametric study. The MAPE value decreased initially as the block size value increased. However, it increased again when the block size exceeded 300. A similar trend was observed for the variation of C value. The MAPE value decreased steadily until C value reached 60. Then, it increased rapidly. Therefore, we suggest that the optimal values for this study are block size = 301 and C= 72.

Figure 8 and Table 2 show the comparison of soil gradation between computer vision (COV) and sieve analysis methods. The mean absolute percent error (MAPE) of coarse-grained soil was low (6.4%) and satisfactory compared to sieve analysis. However, the MAPE of fine-grained soil was high (62 %). This could be due to the insufficient resolution of the camera. The authors suggest that increasing the camera resolution would reduce the error for fine-grained soil. Another possible reason for the underestimation of COV results compared to sieve analysis is the breakage of soil particles during sieving, which creates smaller particle sizes. The detection of particles by sieve analysis is smaller than the non-breakage particles by COV. Further research is needed to find a way to obtain the gradation of soil particles with small sizes, less than 0.4 mm.

This research showed the potential of using COV method to determine soil gradation and classification. The gradation could be obtained in real time in the field by developing a mobile phone or web application that performs cloud-based analysis. The soil classification could be derived from the gradation as well. The proposed method could greatly benefit the quality control of fill in compaction applications with large amounts of fill. The engineer could approve or reject the fill on site without sending soil samples to the laboratory. However, the main challenge of using COV is how to detect the size of fine-grained soil, especially clay particles smaller than 0.076 mm. Further research is needed to calibrate COV to measure the gradation of the wide range of soil particle size.

## 5. CONCLUSIONS

This study proposed a novel computer vision method that uses AruCo target to estimate the particle size distribution of soil from images. The method consists of four steps: noise reduction, adaptive thresholding, particle size detection and gradation calculation. The AruCo target enables the calibration of the pixel size and the actual size of soil particles. The method demonstrates a high accuracy for coarse-grained soil with a mean absolute percent error of 6 % compared to sieve analysis. However, the method has a low accuracy for fine-grained soil with a mean absolute percent error of 63 %. Future work can focus on improving the performance of the method for fine-grained soil and testing it on different types of soil samples.